\documentclass[
]{ceurart}

\usepackage{listings}
\usepackage{tcolorbox}
\begin{document}


\conference{}
\title{Knowledge Graphs of Driving Scenes to Empower the Emerging Capabilities of Neurosymbolic AI}
\author[1]{Ruwan Wickramarachchi}[%
orcid=0000-0002-0877-7063,
email=ruwan@email.sc.edu]
\fnmark[1]

\author[2]{Cory Henson}[%
orcid=0000-0001-7116-9338,
email=cory.henson@us.bosch.com]

\author[1]{Amit Sheth}[%
orcid=0000-0001-7116-9338,
email=amit@sc.edu]

\address[1]{AI Institute, University of South Carolina, Columbia, SC, USA.}

\address[2]{Bosch Center for Artificial Intelligence, Pittsburgh, PA, USA.}



\begin{abstract}
In the era of Generative AI, Neurosymbolic AI is emerging as a powerful approach for tasks spanning from perception to cognition. The use of Neurosymbolic AI has been shown to achieve enhanced capabilities, including improved grounding, alignment, explainability, and reliability. However, due to its nascent stage, there is a lack of widely available real-world benchmark datasets tailored to Neurosymbolic AI tasks. To address this gap and support the evaluation of current and future methods, we introduce DSceneKG --  a suite of knowledge graphs of driving scenes built from real-world, high-quality scenes from multiple open autonomous driving datasets. In this article, we detail the construction process of DSceneKG and highlight its application in seven different tasks. DSceneKG is publicly accessible at: \url{https://github.com/ruwantw/DSceneKG}
\end{abstract}

\maketitle
\section{Introduction}

Integrating intelligent behavior into AI systems requires both perception, processing raw sensor data, and cognition, using background knowledge for tasks like reasoning, planning, and decision-making \cite{sheth2023neurosymbolic}. Knowledge graphs play a crucial role in explicitly representing this background knowledge and enabling AI systems to perform cognitive tasks more effectively. Neural networks, while proficient in pattern recognition, often lack these explicit representations, limiting their ability to perform reliable reasoning.

Neurosymbolic AI aims to overcome this limitation by combining symbolic knowledge representations (e.g., knowledge graphs, ontologies, logical rules) with sub-symbolic AI techniques, such as machine learning and deep learning. Recently, this approach has shown promise in improving reliability, explainability, and performance in handling tasks that demand higher levels of perceptual and cognitive abilities\cite{oltramari2023enabling, gaur2024building}.  However, evaluating such neurosymbolic AI capabilities is often constrained by the use of benchmark datasets that do not reflect the complexities of real-world scenarios, thereby limiting their practical relevance.

A good example of this challenge is provided by knowledge graph completion (KGC), a key problem in knowledge representation and reasoning. Various link prediction (LP) methods have been developed to handle the inherent incompleteness of knowledge graphs by predicting new links in the graph in order to fill in the gaps. These methods are primarily evaluated on standard benchmark datasets like Freebase\cite{bollacker2008freebase} and WordNet\cite{miller1995wordnet}. While such benchmark datasets offer a standardized platform for evaluating LP methods, they do not always accurately capture the complexities of real-world industrial applications. Industries such as automotive, manufacturing, healthcare, and finance are creating large-scale industrial KGs to represent domain-specific data and knowledge. For instance, in the automotive industry, there is a growing demand for large-scale knowledge graphs to represent multi-modal driving scene data from various sensors and cameras, conforming to domain-specific ontologies developed by subject matter experts (SMEs). The reliance on benchmark datasets for evaluating LP methods raises concerns about their applicability to industrial KGs. Real-world KGs differ significantly from benchmark datasets in terms of structure, modality, conformance to ontology, in/out degree, cardinality, etc. Industrial KGs often involve multimodal data, including text, images, and sensor data, and exhibit a higher degree of heterogeneity. This discrepancy highlights the need for more representative benchmarks that can better support the development and evaluation of neurosymbolic AI methods for use in real-world industrial settings.

To address these challenges, we introduce DSceneKG, a suite of knowledge graphs representing real-world driving scenes sourced from multiple autonomous driving datasets. DSceneKG captures a broad spectrum of driving scenarios, including urban and rural environments, various weather conditions, and different traffic situations. By providing a rich symbolic representation of multi-modal data derived from LiDAR, cameras, and GPS sensors, DSceneKG serves as a valuable resource for advancing neurosymbolic AI methods, offering a more realistic and practical benchmark. We will demonstrate the applicability of DSceneKG in developing Neurosymbolic AI solutions for seven tasks: entity prediction, scene clustering/ typing, scene similarity, cross-modal retrieval, root-cause analysis, semantic search, and knowledge completion and augmentation. 

\section{DSceneKG: Driving Scenes Knowledge Graph}
To address these challenges, we introduce DSceneKG, a suite of KGs developed to represent real-world driving data from multiple autonomous driving datasets. DSceneKG captures a wide range of driving scenarios, including urban and rural environments, different weather conditions, and various traffic situations. The data is sourced from several benchmark datasets for AD containing heterogeneous data from LiDAR, cameras, and GPS sensors, providing a rich and multi-modal dataset for research and development in autonomous driving.

\subsection {The Development of DSceneKG}
The DSceneKG primarily contains two main components: (1) The Driving Scenes Ontology (DSO) to represent the formal structure and semantics of scenes, and (2) the generation of a KG based on an existing AD dataset to instantiate the real-world objects and events. Next, we will succinctly describe the development of these two components.

\begin{tcolorbox}[colback=blue!5!white,colframe=blue!75!black, title= Availability of Open Autonomous Driving Datasets]
In recent years, the autonomous driving domain has seen the introduction of several benchmark datasets, including PandaSet (\url{https://pandaset.org/}), NuScenes (\url{https://www.nuscenes.org/}), Waymo Open Dataset (\url{https://waymo.com/open/}), and KITTI (\url{https://www.cvlibs.net/datasets/kitti/}). These datasets provide raw, multimodal sensor data from cameras, LiDAR, and RADAR, along with high-quality annotations. For example, PandaSet, an open-source dataset from Hesai and Scale, features complex driving scenarios from San Francisco and El Camino Real, capturing 103 driving sequences of 8 seconds each. It includes 48K camera images and 16K LiDAR sweeps, with annotations for bounding boxes and semantic segmentations of 38 objects and event categories. NuScenes, by Motional, comprises 1000 driving sequences of 20 seconds each from routes in Boston and Singapore, offering a diverse set of scenes across different continents and conditions. The dataset includes 1.4M camera images, 390K LiDAR sweeps, and 1.4M RADAR sweeps, with annotations that include 3D bounding boxes and object-level attributes. Similarly, the Lyft dataset follows a structure akin to NuScenes, containing 180 driving segments with 22,680 samples. However, it has fewer annotation categories, with only 9 object and event categories compared to NuScenes' 31. These datasets support evaluating tasks such as object detection, segmentation, tracking, behavior/trajectory prediction, and decision-making.
\end{tcolorbox}

\subsubsection {Driving Scenes Ontology}
The Driving Scene Ontology (DSO)\cite{wickramarachchi2021knowledge} provides a formal semantic structure, developed in the Web Ontology Language (OWL) (\url{https://www.w3.org/OWL/}), to represent driving scene information. DSO is designed to be dataset-agnostic and can describe scenes from any autonomous driving dataset. It distinguishes between two types of scenes: \texttt{sequence scene}, representing a sequence over time and space, and \texttt{frame scene}, representing a specific moment in time and space. Temporal information is encoded using specific time instant properties, while spatial information is captured through location names, geographic coordinates, and addresses. The ontology also categorizes entities into objects and events, links them with scenes in which they are observed, and defines relationships for the interactions between objects and events. DSO aims to standardize scene representation, enhancing the understanding and analysis of driving scenarios across different datasets.

\subsubsection {Instantiating a Driving Scene KG from a Public Dataset}

\begin{figure}[!ht]
    \centering
    \includegraphics[width=\linewidth]{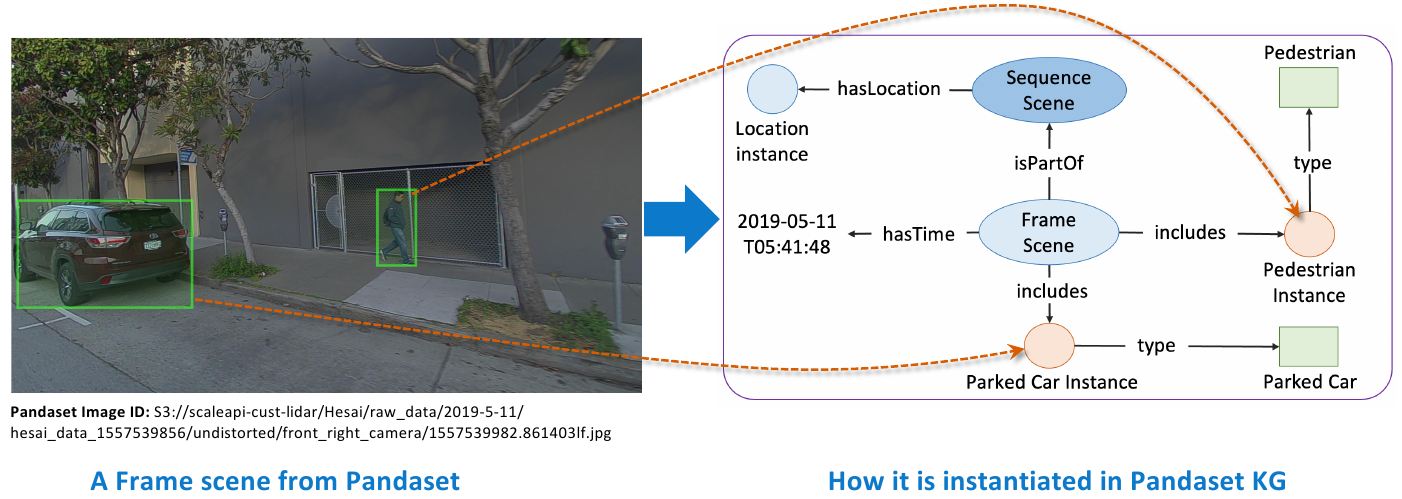}
    \caption{A driving scene from an AD dataset is instantiated as a subgraph in DSceneKG.}
    \label{fig:dscene-kg-inst}
\end{figure}

A KG of driving scenes can be constructed by converting scene data from autonomous driving datasets into Resource Description Framework (RDF)\footnote{https://www.w3.org/RDF/} format, conforming to the Driving Scene Ontology. First, the relevant scene data are extracted using a Software Development Kit (SDK) native to the dataset (e.g., NuScenes-Devkit\footnote{https://github.com/nutonomy/nuscenes-devkit} and Pandaset-Devkit\footnote{https://github.com/scaleapi/pandaset-devkit}). The data are then transformed into RDF using the RDFLib\footnote{https://rdflib.readthedocs.io/en/stable/} Python library.  Each entity is associated only with the specific frame scene in which it appears, ensuring an accurate representation of the scene data. For example, Figure \ref{fig:dscene-kg-inst} illustrates how a real-world driving scene from Pandset can be instantiated as a subgraph in DSceneKG. Table \ref{tab:kg-stats} summarizes the statistics of the resultant KGs constructed from NuScenes, Pandaset, and Lyft. 

\begin{table}[h]
\centering
\scriptsize
\begin{tabular}{c|ll|lll}
\toprule
& \textbf{Freebase (FB15k)} & \textbf{WordNet (WN11)} & \textbf{NuScenes} & \textbf{Pandaset} & \textbf{Lyft}  \\
\midrule
\#triples        & 592,212   & 151,441  & 6,296,378   & 3,301,928 & 3,944,516  \\
\#entities       & 14,951    & 40,943   & 2,108,545   & 53,248 & 1,327,255    \\
\#relations      & 1,345     & 18      & 14       & 19 & 13      \\
Avg. in-degree    & 39.6633  & 3.6992  & 3.0353   & 62.1387 & 3.0237 \\
Avg. out-degree   & 39.6633  & 3.6991  & 3.0107   & 63.3269 & 2.9824  \\
Triples/entities & 39.6102  & 3.6988  & 2.9861   & 62.0104 & 2.9719\\ 
\bottomrule
\end{tabular}
\caption{Descriptive statistics of benchmark KGs vs. real-world KGs developed from the autonomous driving domain.}
\label{tab:kg-stats}
\end{table}

\section{Applications of DSceneKG: Emerging Neurosymbolic AI Capabilities}

\begin{figure}[!ht]
    \centering
    \includegraphics[width=1.0\linewidth]{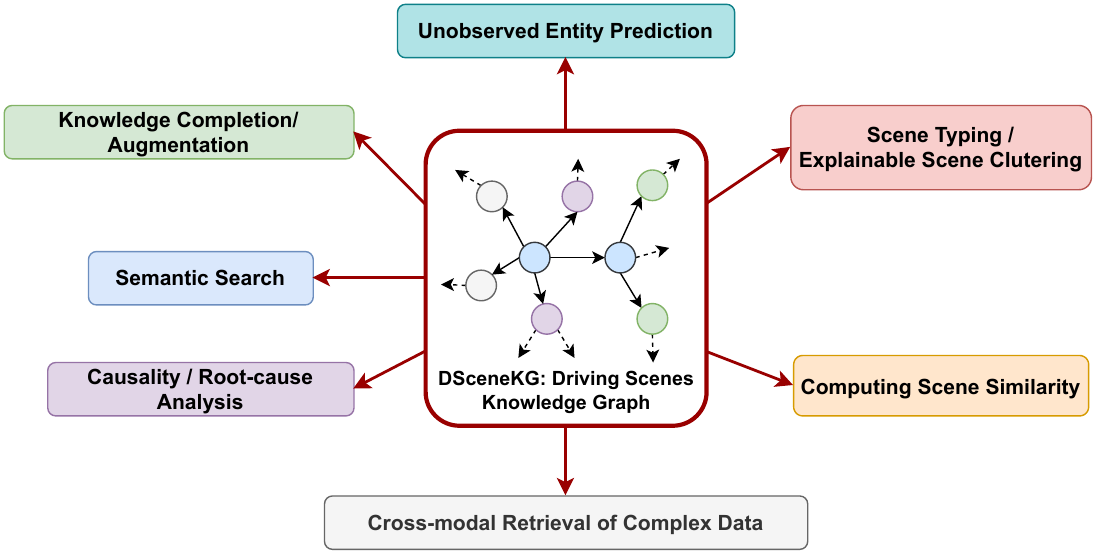}
    \caption{Applications of DSceneKG across seven different problems in AI}
    \label{fig:dscene-kg-applications}
\end{figure}

DSceneKG has significant potential for both industrial and academic applications. In this section, we showcase seven Neurosymbolic solutions that use DSceneKG as the benchmark dataset for evaluation (see Figure \ref{fig:dscene-kg-applications}).  

\subsection{Machine Perception}

Machine perception enables autonomous systems to operate effectively in dynamic environments. As these systems become increasingly integrated into daily life, across domains like transportation, manufacturing, and healthcare, their ability to sense and adapt to changing conditions is vital. Autonomous driving, for instance, serves as a key test-bed for solving complex AI problems. Scene understanding is a critical component, which involves comprehending and interpreting various aspects of a scene, such as the detection, recognition, and localization of objects and events. We will showcase solutions for three scene understanding tasks built around DSceneKG.

\subsubsection{Knowledge-based Entity Prediction}
Knowledge-based entity prediction (KEP) involves predicting the presence of potentially unrecognized or unobserved entities in a scene using current and background knowledge represented in a knowledge graph. The goal is to leverage an expressive KG to provide high-level semantic cues for identifying entities that are not explicitly recognized by traditional perception systems. For example, if an autonomous vehicle detects a ball on the road in a residential area, KEP would help predict the likely presence of a child nearby, considering knowledge about the context and relationships between objects, like children playing with balls. To address this issue, \cite{wickramarachchi2021knowledge} leverages the holistic and expressive scene representation in DSceneKG to build a link prediction-based solution for KEP. They demonstrate the effectiveness of this approach by showing that the missing entities may be predicted with high precision (0.87 Hits@1) while significantly outperforming the non-semantic and rule-based baselines.

\subsubsection{Explainable Scene Clustering/ Typing}
KGs can help to explore sets of interrelated entities and discover meaningful patterns by clustering entities into informative subsets. For example, in the context of autonomous driving, KGs could organize scene types --- e.g., scenes around a school zone, scenes around an accident --- and entities, such as vehicle types, road signs, traffic conditions, and pedestrian behaviors into clusters, aiding users in understanding the relationships and relevance of these entities. However, clustering alone is insufficient. The nature of each cluster must also be understandable. Simple labels like ``vehicle'' or ``pedestrian'' are too broad, while fine-grained types can overwhelm users with too many labels. The challenge is to find a balance, using KGs to create clusters and generate concise, user-comprehensible labels that accurately represent the clusters' contents. Therefore, explainable scene clustering involves clustering entities into semantically similar groups and providing clear, concise explanations for these clusters based on their relationships within the KG. \cite{nag2021towards} developed a solution based on DSceneKG to address this problem. They further showcase how DSceneKG can be further enriched with knowledge about \textit{commonsense relations} to improve the context understanding and provide better explanations.

\subsubsection{Computing Semantic Similarity}
Computing scene similarity involves determining how \textit{alike} two scenes are based on certain features or characteristics. In autonomous driving, addressing this problem by only considering visual characteristics is problematic as driving scenes can be visually dissimilar but semantically similar. For example, consider the scenes recorded from two physical locations where a vehicle turns left from a roundabout. The visual information can be quite dissimilar; however, the high-level action is the same. \cite{wickramarachchi2020evaluation} proposes a solution based on DSceneKG where they first transform the DSceneKG into embedding vectors and compute the cosine similarity between the vectors of scene pairs to identify those with the highest similarity scores. Notably, this method could detect similarities even when scenes were not visually alike, focusing instead on shared, high-level semantic characteristics.

\subsection{Knowledge Completion and Augmentation}
Knowledge graph completion refers to the task of completing a graph with missing information, i.e. filling in the gaps. Different types of knowledge may need to be completed, such as missing relations, entities, and high-level entity-type information of instances that are currently typed to only their granular types. DSceneKG will enable real-world evaluations of the current and future knowledge completion methods for the above-mentioned tasks. Additionally, DSceneKG facilitates the evaluation of methods designed to complete knowledge specific to driving scenes. For example, \cite{wick2023_clue} proposes a context-based approach for labeling unobserved entities in DSceneKG. The scene nodes in DSceneKG can then be augmented with these newly obtained labels for entities that may have gone unobserved or unlabeled in the original dataset.

\subsection{Semantic Search}
For tasks requiring semantic search over multimodal data, DSceneKG can be utilized in two primary ways. First, DSceneKG can be queried directly using SPARQL, as all visual scene elements and metadata are structured within the graph, enabling efficient search and inference tasks. Second, vector search can be performed over the learned embeddings by leveraging vector databases, which efficiently store high-dimensional vectors. This allows for fast and accurate similarity searches based on vector similarity. By storing knowledge graph embeddings (KGEs) generated through state-of-the-art KGE methods, search operations can run directly over the learned vector space. For instance, \cite{Henson2019UsingAK} discusses the benefits of using scene KGs in an enterprise setting to enable efficient ``scene-based" search over heterogeneous information. 

\subsection{Causality}
Causality is often studied using frameworks like Causal Bayesian Networks (CBNs), which represent variables and their causal relationships as directed acyclic graphs (DAGs). CBNs allow for reasoning about cause and effect by leveraging probabilistic relationships between variables, enabling estimations about how changes in one variable can influence others. Recently, there has been an interest in improving the representation of causality with knowledge graphs where domain knowledge graphs that represent observation data are enriched with information from causal Bayesian networks (CBN) to enhance causal inference and explainability. For example, \cite{jaimini2022causalkg} showcases how DSceneKG can be enriched with causal information to create a causal knowledge graph that enables counterfactual and intervention reasoning to understand the behaviors of scene entities.

\subsection{Cross-modal Retrieval of Complex Data}
Cross-modal retrieval aims to retrieve relevant information in one modality based on a query in another. For example, in autonomous driving, cross-modal retrieval aims at retrieving bird's-eye-view (BEV) scene representations (e.g., image/video) from textual input or instructions. In such cases, using a global semantic structure is essential to provide semantic relationships between entities like objects, features, and movements. Building upon this idea, \cite{wei2024bev} proposes a novel BEV retrieval method that uses the DSceneKG as a source of associative embeddings, enriching the representation of input text by embedding related autonomous driving keywords. These keyword embeddings are integrated with language models, enabling better alignment between the text input and the bird's eye view (BEV) features extracted from visual data. This fusion of knowledge graph embeddings and text descriptions improves the retrieval of BEV representations by providing a more structured and semantically relevant context.

\section{Conclusions}
We introduce a suite of driving scenes knowledge graphs, DSceneKG, designed to benchmark the emerging capabilities of Neurosymbolic AI. Built from real-world, open-domain datasets, DSceneKG integrates multimodal data from diverse driving scenes across various continents and environmental conditions. We outline the process of constructing DSceneKG and demonstrate its application across seven different tasks.

\section{Acknowledgments}
This research was supported in part by NSF grants \#2133842 and \#2119654.  The opinions expressed are those of the authors and do not necessarily reflect the views of the NSF.

\section*{Authors}

\noindent \textbf{Ruwan Wickramarachchi} is a Ph.D. candidate at the AI Institute, University of South Carolina. His research focuses on introducing a Neurosymbolic approach to scene understanding to improve machine perception and context understanding in autonomous systems (e.g., autonomous driving and smart manufacturing). Contact him at: \url{ruwan@email.sc.edu} \\

\noindent \textbf{Cory Henson} is a lead research scientist at the Bosch Center for Artificial Intelligence in Pittsburgh, USA. Contact him at: \url{cory.henson@us.bosch.com} \\

\noindent \textbf{Amit Sheth} is the NCR Chair, and a professor; he founded the university-wide AI Institute of South Carolina (AIISC) in 2019. He received the 2023 IEEE-CS Wallace McDowell award and is a fellow of IEEE, AAAI, AAIA, AAAS, and ACM. Contact him at: \url{amit@sc.edu}

\bibliography{references}

\end{document}